\documentclass[10pt,twocolumn,letterpaper]{article}

\usepackage[pagenumbers]{wacv} %

\usepackage{tikz}
\usepackage{pgfplots}
\pgfplotsset{compat=1.18}

\definecolor{wacvblue}{rgb}{0.21,0.49,0.74}
\usepackage[pagebackref,breaklinks,colorlinks,allcolors=wacvblue]{hyperref}
\usepackage{algorithm}
\usepackage{algpseudocode}
\def\wacvPaperID{2836} %
\def\confName{WACV}
\def\confYear{2027}

\title{Domain Adaptation of SAM 2.1 for Building-Component Segmentation in Image-Based 3D Reconstruction}

\author{
Ehsan Agha Ebrahimi\\
LSU Dept. Const. Mgmt.\\
{\tt\small eaghae1@lsu.edu}
\and
Yong-Cheol Lee\\
LSU Dept. Const. Mgmt.\\
{\tt\small yclee@lsu.edu}
\and
Keith G. Mills\\
LSU Div. CS\&E \\
{\tt\small keith.mills@lsu.edu}
}

\begin{document}
\maketitle

\begin{abstract}
3D reconstruction of building components using image-based scanning requires appropriate segmentation of the objects in the 2D building images. In order to benefit from consistent grouping of the 2D masks in 3D environments, vision foundation models such as the Segment Anything Model have been utilized by Group Anything with Radiance Fields (GARField) to perform 3D segmentation. This study fine-tuned SAM~2.1 on the HBD dataset, consisting of 2,679 annotated images from 42 buildings, to enhance multi-class segmentation of 2D building images, covering 11 classes (Wall, Window, Ceiling, Door, Column, Beam, Floor, Opening, Curtain Wall, Roof, and Lift). The model was evaluated using per-class IoU, Dice, and mIoU metrics, and the results show an improvement in mIoU from 49.9\% to 69.6\%, indicating a meaningful improvement in class segmentation and providing a more accurate model for distilling the masks in 3D. SegFormer was then used as a baseline, generating per-class masks for comparison with SAM~2.1, which indicates improvement in some object classes with larger surface area. The masks in the building images were grouped in the neural radiance field, and the pretrained SAM~2.1 was compared with the fine-tuned SAM~2.1 in the newly rendered images. This comparison shows that fine-tuning improves the grouping of building components at a user-selected grouping scale. This advances the application of Group Anything with Radiance Fields (GARField) in 3D, enabling better segmentation of building structures for 3D auto-reconstruction.
\end{abstract}
    
\section{Introduction}
\label{sec:intro}
\begin{figure}[t]
  \centering
  \includegraphics[width=\linewidth]{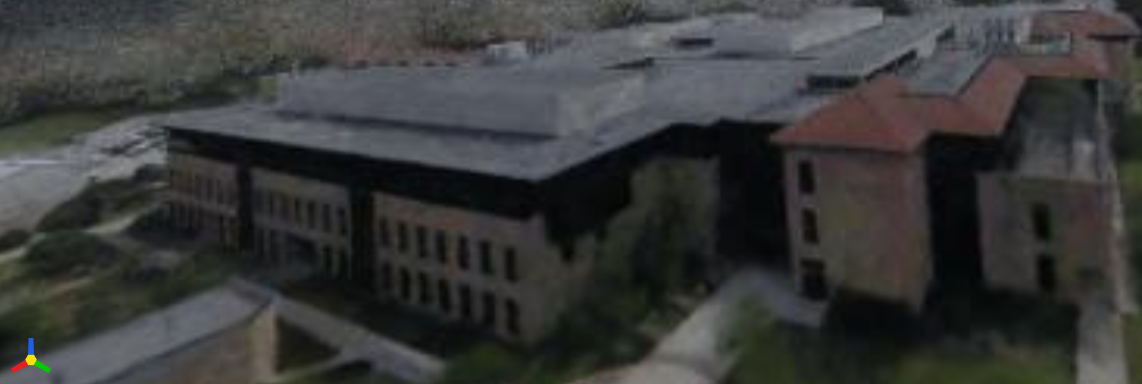}\\[2pt]
  \includegraphics[width=\linewidth]{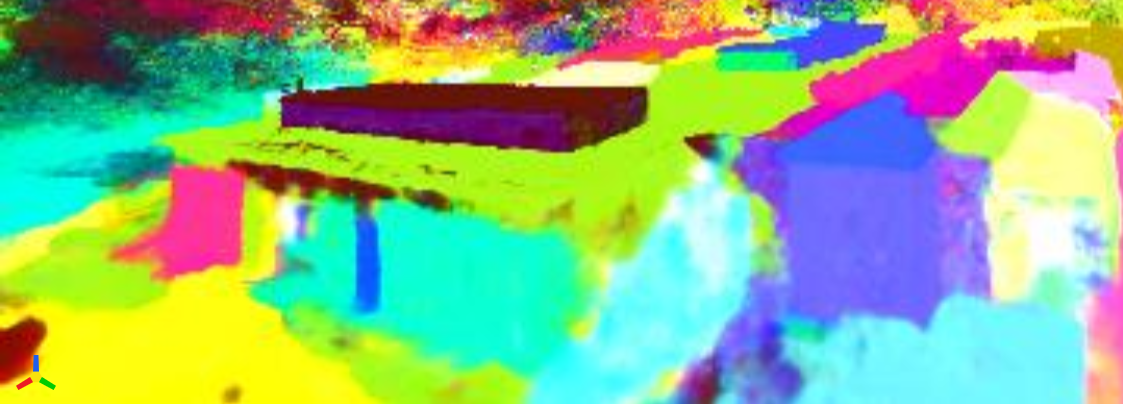}
  \caption{Patrick F. Taylor Hall reconstructed in 3D from drone images.
  Top: rendered photo view. Bottom: the same view segmented into building
  components with our fine-tuned SAM~2.1 and grouped hierarchically in 3D
  by Group Anything with Radiance Fields (GARField)~\cite{garfield2024},
  with each 3D group shown in a distinct color.}
  \label{fig:pft_teaser}
\end{figure}
3D scene reconstruction using 2D images is one of the areas that has several downstream applications, such as capturing the reality and creating a digital twin of the real world.  In this regard, semantic segmentation of objects in 3D by 2D images has always been a matter of concern in computer vision, and several studies, such as SA3d \cite{cen2025segment} and OpenNeRF\cite{engelmann2024opennerfopenset3d}, tried to address this issue by either using deep learning models or distilling the 2D masks in 3D. Although methods such as photogrammetry and NeRF can reconstruct geometry, they do not provide reliable semantic labels for building components \cite{mildenhall2020nerf}. One of the recent studies for 3D segmentation of the objects in 3D, benefits from Neural Radiance Field (NeRF), which makes the possibility to create a scale-conditioned 3D affinity field using 3D feature vectors for grouping the mask by different scales in the 3D environment \cite{garfield2024}. 

This study aims to improve object-oriented 3D reconstruction of buildings from 2D images by
adapting a promptable vision foundation model to the building domain so that building
components are segmented and grouped consistently in 3D. The focus is on identifying the
main building components in 2D images, such as walls, windows, doors, and roofs, which
are essential for converting image-based 3D models into structured representations. Existing segmentation models are not adapted to the building domain, and they are
designed to produce masks for a predefined set of object classes. Models such as
YOLOv8~\cite{yolov8_ultralytics}, SAM~3~\cite{carion2025sam3segmentconcepts} (given an open-vocabulary text prompt), and
SegFormer~\cite{xie2021segformer} segment a scene into masks for specific object
categories. However, to maintain 3D consistency during
reconstruction, masks must be generated automatically for the entire scene rather than
only for a fixed set of classes.
Among current models, the vision foundation model SAM~2.1~\cite{ravi2024sam2} can
automatically generate masks for every part of a scene, which makes it a suitable source
of masks for 3D grouping with Group Anything with Radiance Fields
(GARField)~\cite{garfield2024}.

A key property we exploit is the scale of the masks, which GARField uses in a hierarchical manner during training and consolidates inconsistent 2D masks. The building masks exist at different scales: a mask may cover a complete window, while other masks cover parts of windows or doors. Training with this hierarchy of mask scales leads to more consistent building segmentation when the masks are grouped together. 

The Segment Anything Model (SAM) is a vision foundation model developed by Meta AI for
promptable image segmentation~\cite{kirillov2023segany}. Subsequently, SAM~2.1 was
introduced as a unified model for image and video segmentation that uses point, box, and mask prompts on individual frames to define the spatial extent of an object and segment it spatio-temporally~\cite{ravi2024sam2}. Its architecture consists of an image encoder (a vision transformer), a prompt encoder, and a mask decoder. SAM~2.1 was trained on the SA-V dataset, which comprises 50.9K videos with 642.6K masklets~\cite{ravi2024sam2}. Although several methods exist for adapting the model, such as adapter-based fine-tuning~\cite{chen2024sam2adapterevaluatingadapting} that modifies the weights of the
upper layers, the fine-tuning approach used in this paper is the one provided by Meta AI to fine-tune SAM~2.1 on a custom dataset \cite{chen2024sam2adapterevaluatingadapting}.

\begin{figure*}[t]
  \vskip 0.4in
  \begin{center}
    \centerline{\includegraphics[width=\textwidth]{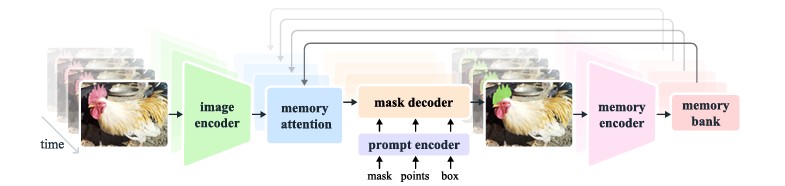}}
    \caption{
      The SAM 2 architecture (figure adopted from \citet{ravi2024sam2}).
    }
    \label{fig:sam2_architecture}
  \end{center}
\end{figure*}

First, we used a custom dataset created by \citet{WONG2024105558}, consisting of interior
and exterior images of buildings, for building-component segmentation. Although many
datasets exist for building image segmentation, this dataset was chosen because it was
created specifically for 3D reconstruction of building components. Its annotations cover
most of the facade and interior, and it is already split into training, validation, and
test sets.

Second, both SAM~2.1 and the baseline, SegFormer, were fine-tuned on this dataset, and
their performance was evaluated using segmentation metrics. The results indicate that
SAM~2.1 improves segmentation for some discrete object types. However, for large-surface
objects such as walls, SegFormer outperforms SAM~2.1 in those classes.

Third, the comparison was carried out by rendering existing and novel views of the
building. The experimental results show that the fine-tuned model performs better in the
hierarchical grouping of the building facade when different grouping scales are applied.
The groups are visible across different views, and the windows, doors, and walls of the
building are grouped together as part of the facade when the scale is large enough to
merge closely related affinity vectors.

\section{Related Work}

Semantic and instance segmentation of building images has been widely studied for facade
parsing and building-component recognition. Early work such as DeepFacade fine-tuned an
ImageNet-pretrained network to parse facades into components such as windows, walls, and
doors~\cite{ijcai2017p320}. More recent work adapts stronger instance-segmentation models
to the building domain; for example, \citet{LU2023113275} trained a SOLOv2 model on a
large dataset of interior and exterior building images to improve building-component
parsing. That study focused on window recognition and aimed to improve the efficiency of
window detection under occlusion and reflection. Other studies detect facade elements from
images using prior structural knowledge~\cite{ZHANG2022103068}; that work combined
structure from motion with a multimodal semantic feature map to perform hierarchical
clustering of the facade. Further studies use deep learning models to extract specific
components, such as the method of \citet{isprs-archives-XLI-B3-709-2016}, which detects windows in 2D
images.

In recent years, 3D modeling with semantic segmentation of building components has
received considerable attention. For instance, \citet{Salehitangrizi31122024} used mobile
scanning images to capture multiple views of a building and applied an R-CNN to detect
doors and windows on the facades. Other studies segment directly in 3D; for example,
\citet{10208594} used a ray-casting and Bayesian-network method rather than transferring
the segmentation from 2D to 3D. In addition, several studies apply deep learning to point
clouds for semantic segmentation of facades and building components; \citet{10943351} is
one recent example that learns per-class semantics for point clouds. Although these studies
provide suitable models for segmentation in 2D images or in 3D point clouds, the
reconstruction of building components from 2D segmentation remains understudied. The main
challenge lies in distilling the 2D segments into the 3D environment. To address this, this
paper combines an existing grouping method with the domain adaptation of a vision
foundation model.

\section{Methodology}
\label{sec:formatting}
Our method consists of evaluation of the fine-tuned model to assess the accuracy of the segmentation along with the per-class baseline comparison with the fine-tuned SegFormer to evaluate the performance of the fine-tuned SAM~2.1, visualized in Figure~\ref{fig:sam2_architecture}. 

\subsection{Training Function}
Fine-tuning SAM~2 begins from pretrained weights $\theta_0$ and optimizes them on the target building facade dataset. At each training step, a user prompt $p$ is simulated by sampling a random point or bounding box from the ground truth mask $M_{gt}$. The model predicts a segmentation mask through a forward pass:
\begin{equation}
    M_{pred} = D\bigl(E_{img}(I;\,\theta),\; E_{prompt}(p;\,\theta)\bigr)
\end{equation}
where $E_{img}$ denotes the image encoder, $E_{prompt}$ the prompt encoder, and $D$ the mask decoder. The prediction is evaluated using a composite loss:
\begin{equation}
    \mathcal{L} = 20 \cdot \mathcal{L}_{BCE} + \mathcal{L}_{Dice} + \mathcal{L}_{IoU} + \mathcal{L}_{class}
\end{equation}
which combines per-pixel binary cross-entropy, region overlap, confidence calibration, and object detection terms. Backpropagation computes the gradients $\partial \mathcal{L}/\partial \theta$ via the chain rule, and the AdamW optimizer updates the weights \cite{loshchilov2019decoupledweightdecayregularization}.

\subsection{Evaluation}
We evaluate this work by performing the quantitative semantic evaluation metrics, IoU per class, along with the mean and the dice score per class. We evaluate SAM~2.1 and pre- and after fine-tuning with some portions of HBD dataset and use 349 images to  calculate IoU and Dice score for 5541 instances in these images for each specific class regarding the ground truth masks.  

\begin{figure}[t]
\centering
\includegraphics[width=\linewidth]{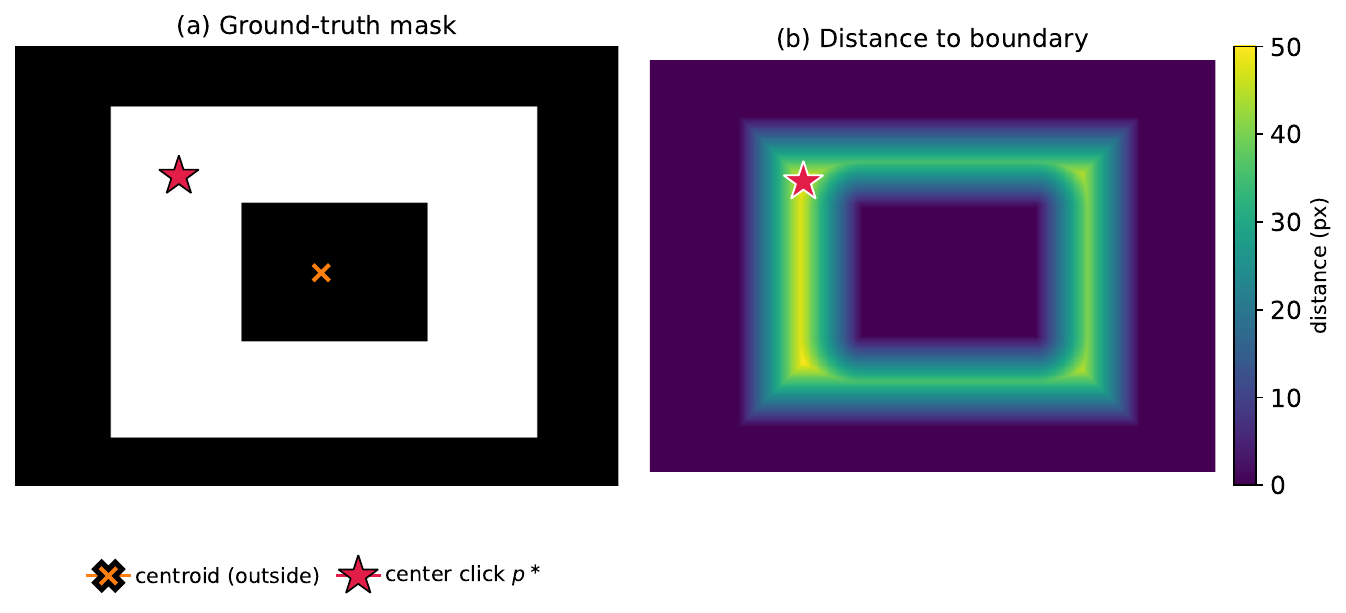}
\caption{Center-click prompt selection. For each ground-truth component mask we
compute the Euclidean distance of every interior pixel to the mask boundary~(b)
and take the pixel of maximum distance, $p^\ast$ (red star) as a single
positive click for SAM~2. This deepest point always lies inside the object,
even for ring-shaped components where the centroid ($\times$) falls outside the
mask~(a).}
\label{fig:center_click}
\end{figure}

To measure segmentation quality under a controlled and reproducible prompt, we
follow the interactive protocol of SAM~2~\cite{ravi2024sam2} and query the model
with a single positive point per object. Since SAM~2 can only generate masks automatically with a box or click prompt and not by the class text prompt, the mask is generated by giving the center of the ground truth masks as a first click prompt. Then the generated mask is evaluated with the corresponding ground truth to evaluate improvement in segmentation after fine-tuning. For each ground-truth component mask
$M$, we select the prompt point $p^\ast$ as the interior pixel that is farthest
from the mask boundary,
\begin{equation}
p^\ast = \arg\max_{p \in M} \; d\!\left(p, \partial M\right),
\label{eq:center}
\end{equation}
where $\partial M$ is the boundary of $M$, $d(\cdot,\cdot)$ is the Euclidean
distance, and the image border is treated as part of the boundary. In practice,
$p^\ast$ is obtained from the Euclidean distance transform of $M$ and
corresponds to the center of the largest disk inscribed in the object. This
``deepest'' point is stable and unambiguous: unlike the centroid it is
guaranteed to lie inside the mask even for concave or ring-shaped components
(Fig.~\ref{fig:center_click}), and unlike a randomly sampled point, it is
deterministic, yielding the same prompt on every run. The selected point is
passed to SAM~2.1 as a single positive click, and the predicted mask is compared
against $M$. Because the prompt is derived from the ground truth, the identical
rule is applied to every model (pre-trained SAM~2.1, fine-tuned SAM~2.1, and all
ablations); the metric therefore isolates per-class mask quality rather than the
model's ability to localize or classify objects. The class label used for
per-class scoring is likewise taken from the ground-truth annotation.

\begin{algorithm}[t]
\caption{Center-click prompt selection and evaluation}
\label{alg:center_click}
\begin{algorithmic}[1]
\Require image $I$; ground-truth pairs $\{(M_k, c_k)\}_{k=1}^{K}$ (mask, class); model $f$
\Ensure per-class IoU and Dice
\State $f.\textsc{SetImage}(I)$ \Comment{encode the image once}
\For{each object $(M_k, c_k)$}
    \State $D \gets \textsc{DistanceTransform}(M_k)$ \Comment{Euclidean distance to $\partial M_k$}
    \State $p^\ast \gets \arg\max_{p} D(p)$ \Comment{deepest interior pixel = center click}
    \State $\hat{M}_k \gets f.\textsc{Predict}(\text{point}{=}p^\ast,\ \text{label}{=}{+}1)$ \Comment{one positive click}
    \State $\text{IoU}_k \gets \dfrac{|\hat{M}_k \cap M_k|}{|\hat{M}_k \cup M_k|}, \quad
            \text{Dice}_k \gets \dfrac{2\,|\hat{M}_k \cap M_k|}{|\hat{M}_k| + |M_k|}$
    \State append $\text{IoU}_k,\ \text{Dice}_k$ to class $c_k$
\EndFor
\State \Return per-class means, and macro-averages mIoU, mDice
\end{algorithmic}
\end{algorithm}

\subsection{Baseline Comparison}

The fine-tuned SAM~2.1 is compared with two baselines on the same test split of the
dataset: (i)~the pre-trained SAM~2.1 (Hiera-L) checkpoint and (ii)~SegFormer
\cite{xie2021segformer}, a transformer-based semantic segmentation network,
fine-tuned on the same training split under matched settings. The three models
produce different outputs and are in different forms, which we reconcile as follows.
SegFormer is a Transformer-based model for semantic segmentation, which assigns a
class to every pixel and hence assigns one mask per class directly
\cite{xie2021segformer}. SAM~2.1, in contrast, is a promptable, class-agnostic model
that returns a single mask given a prompt but no class label and is not capable of
producing per-class masks itself, unlike SAM~3~\cite{carion2025sam3segmentconcepts}
whose text/concept prompts enable open-vocabulary per-class segmentation.

To train SegFormer, which requires a single class label per pixel, we convert the
per-object instance masks into semantic label maps: for each image we render one
single-channel map in which every pixel holds its component class (0 for background).
Where annotations overlap (for example a window within a wall), specific fixtures
take priority over the enclosing surface, so each pixel receives its most specific
class. We fine-tune SegFormer with the backbone of MiT-B3 encoder pre-trained on ImageNet at
$512\times512$ resolution using the HuggingFace \texttt{Trainer}~\cite{wolf2020transformers,hf_trainer_docs},
with AdamW \cite{loshchilov2019decoupledweightdecayregularization} (learning rate $6\times10^{-5}$), a batch size of 4, and random horizontal
flipping, for up to 80 epochs; we evaluate on the validation split after every epoch
and keep the checkpoint with the lowest validation loss.

For evaluation, SegFormer is run fully automatically---it predicts both the mask and
the class with no prompt---and is scored per pixel. SAM~2.1 (pre-trained and
fine-tuned) is prompted with a single center click derived from the ground-truth mask
and takes its class from the ground-truth annotation, and is scored per object
(Sec.~\ref{sec:eval}). These two protocols are not directly comparable, and we make
the difference explicit: they differ in (i)~prompting---SAM receives
ground-truth-derived guidance whereas SegFormer does not---and (ii)~aggregation---per
object for SAM versus per pixel for SegFormer. We therefore treat SegFormer as a
reference for a fully-automatic semantic segmenter rather than a strict head-to-head
competitor, and we draw quantitative conclusions only from the matched
pre-trained-vs-fine-tuned SAM~2.1 comparison, which uses identical prompts and
scoring. A further limitation is that flattening overlapping annotations into a
single label per pixel discards the hierarchical, multi-scale structure of the
dataset, which a semantic segmenter cannot represent; under this flattening a small
fraction of images (25 of 3{,}378) lose one class to unavoidable overlaps.

\subsection{GARField Integration and Evaluation}

To evaluate the effect of our fine-tuning beyond 2D segmentation, we integrate the segmentation model into GARField~\cite{garfield2024} (Group Anything with Radiance Fields), which lifts 2D group masks into a 3D, scale-conditioned affinity field so that building components can be grouped consistently across views and at multiple scales. GARField supervises this field with 2D masks from an automatic mask generator; we use SAM~2.1 as this generator and compare two conditions: the pretrained SAM~2.1 and our fine-tuned SAM~2.1. In both conditions the masks are produced automatically (a grid of point prompts, no ground-truth guidance), unlike our prompted 2D evaluation, since no ground truth is available at this stage.

The comparison is designed to isolate the effect of the fine-tuning. We reconstruct the scene once and train GARField twice under identical settings---same images, camera poses, radiance field, grouping scales, and training schedule---so that the only difference between the two runs is the SAM~2.1 checkpoint used to generate the masks. We then query the trained grouping field at a fixed scale and render the grouping from the captured viewpoints, comparing the pretrained and fine-tuned results. We note that the hierarchical grouping mechanism itself is GARField's contribution, which we adopt and cite; our claim is only that fine-tuning the mask generator on building components improves the quality of that grouping in this domain.

We assess the grouping qualitatively and render the grouping field as a per-point feature map reduced to RGB by PCA and place it beside the rendered image, so that a cleaner, more complete grouping of components (for example, each window as a single region) is directly visible.

\section{Results and Discussion}
The results evaluate the effect of fine-tuning in terms of training loss, per-class segmentation accuracy, comparison with SegFormer, and qualitative 3D grouping at different GARField grouping scales.
\subsection{Quantitative Evaluation}\label{sec:grouping_quant}
\subsubsection{Loss}
To fine-tune SAM~2.1, we used the HBD dataset \cite{WONG2024105558}, which contains 2,679 annotated images of 42 buildings. The dataset includes interior and exterior images with annotations provided in JSON format. We converted these annotations into PNG masks, generating a separate mask for each object instance. We fine-tuned the SAM~2.1 Hiera-Large checkpoint (sam~2.1-hiera-large.pt) using the multi-mask training configuration. Because fine-tuning is computationally intensive, training was performed on LSU’s SuperMike3 HPC cluster using one compute node with four NVIDIA A100 GPUs. Training ran for 40 epochs (1,340 steps per epoch; 53,600 total steps) and required 18 hours and 5 minutes. The total loss decreased by 55.9\%, from 1.864 to a minimum of 0.823 at epoch 31.

\begin{figure}[htbp]
    \centering
    \begin{tikzpicture}
        \begin{axis}[
            width=0.95\columnwidth,
            height=0.55\textwidth,
            xlabel={Epoch},
            ylabel={Loss},
            title={SAM~2 Fine-tuning Training Loss Convergence},
            legend style={
                at={(0.97,0.97)},
                anchor=north east,
                font=\small,
                cells={anchor=west},
                draw=black,
                fill=white,
                fill opacity=0.8,
            },
            grid=major,
            grid style={dashed, gray!30},
            xmin=0, xmax=39,
            ymin=0, ymax=2.0,
            xtick={0,5,10,15,20,25,30,35,39},
            tick label style={font=\small},
            label style={font=\small},
            title style={font=\normalsize\bfseries},
        ]

        \addplot[
            color=blue,
            mark=*,
            mark size=1.5pt,
            thick,
            line join=round,
        ] coordinates {
            (0,1.8642) (1,1.4015) (2,1.2601) (3,1.1854) (4,1.1401)
            (5,1.0989) (6,1.0720) (7,1.0099) (8,1.0537) (9,0.9796)
            (10,0.9978) (11,0.9974) (12,1.0263) (13,0.9589) (14,0.9202)
            (15,0.9234) (16,0.9808) (17,0.9322) (18,0.9296) (19,0.9296)
            (20,0.8836) (21,0.8532) (22,0.9166) (23,0.9012) (24,0.8548)
            (25,0.8873) (26,0.8497) (27,0.9028) (28,0.9218) (29,0.8753)
            (30,0.8575) (31,0.8225) (32,0.8761) (33,0.8724) (34,0.8496)
            (35,0.8631) (36,0.8818) (37,0.8373) (38,0.8422) (39,0.8869)
        };
        \addlegendentry{Total Loss}

        \addplot[
            color=red,
            mark=triangle*,
            mark size=1.5pt,
            thick,
            line join=round,
        ] coordinates {
            (0,0.8033) (1,0.6200) (2,0.5660) (3,0.5334) (4,0.5171)
            (5,0.5036) (6,0.4928) (7,0.4570) (8,0.4830) (9,0.4436)
            (10,0.4446) (11,0.4513) (12,0.4693) (13,0.4485) (14,0.4158)
            (15,0.4199) (16,0.4374) (17,0.4229) (18,0.4324) (19,0.4199)
            (20,0.4011) (21,0.3973) (22,0.4214) (23,0.4086) (24,0.3988)
            (25,0.4129) (26,0.3948) (27,0.4098) (28,0.4251) (29,0.4046)
            (30,0.3909) (31,0.3800) (32,0.4056) (33,0.4023) (34,0.3972)
            (35,0.3870) (36,0.4023) (37,0.3815) (38,0.3808) (39,0.3968)
        };
        \addlegendentry{Dice Loss}

        \addplot[
            color=orange,
            mark=square*,
            mark size=1.5pt,
            thick,
            line join=round,
        ] coordinates {
            (0,0.4040) (1,0.3092) (2,0.2859) (3,0.2737) (4,0.2602)
            (5,0.2560) (6,0.2543) (7,0.2378) (8,0.2493) (9,0.2299)
            (10,0.2345) (11,0.2419) (12,0.2481) (13,0.2350) (14,0.2137)
            (15,0.2204) (16,0.2321) (17,0.2249) (18,0.2271) (19,0.2229)
            (20,0.2152) (21,0.2092) (22,0.2272) (23,0.2172) (24,0.2141)
            (25,0.2166) (26,0.2109) (27,0.2185) (28,0.2235) (29,0.2122)
            (30,0.2064) (31,0.2037) (32,0.2166) (33,0.2156) (34,0.2071)
            (35,0.2084) (36,0.2175) (37,0.2071) (38,0.2045) (39,0.2138)
        };
        \addlegendentry{IoU Loss}

        \addplot[
            color=green!60!black,
            mark=diamond*,
            mark size=1.5pt,
            thick,
            line join=round,
        ] coordinates {
            (0,0.6570) (1,0.4723) (2,0.4082) (3,0.3783) (4,0.3627)
            (5,0.3393) (6,0.3249) (7,0.3151) (8,0.3214) (9,0.3061)
            (10,0.3187) (11,0.3042) (12,0.3089) (13,0.2753) (14,0.2907)
            (15,0.2830) (16,0.3112) (17,0.2845) (18,0.2701) (19,0.2869)
            (20,0.2673) (21,0.2468) (22,0.2680) (23,0.2754) (24,0.2418)
            (25,0.2578) (26,0.2441) (27,0.2745) (28,0.2732) (29,0.2584)
            (30,0.2603) (31,0.2388) (32,0.2539) (33,0.2544) (34,0.2453)
            (35,0.2678) (36,0.2620) (37,0.2487) (38,0.2569) (39,0.2763)
        };
        \addlegendentry{Mask Loss ($\times 20$)}

        \draw[dashed, gray, thick] (axis cs:31,0) -- (axis cs:31,0.8225);
        \node[font=\scriptsize, anchor=south] at (axis cs:31,0.82) {Best: 0.8225};

        \end{axis}
    \end{tikzpicture}
    \caption{Training loss convergence during SAM~2 Hiera-Large fine-tuning on the HBD building facade dataset over 40 epochs. The total loss decreased by 55.9\% from 1.864 to a minimum of 0.823 at epoch 31, demonstrating successful adaptation to the architectural domain. The model was trained on 2{,}679 images using 4$\times$ NVIDIA A100 GPUs for 18 hours.}
    \label{fig:sam2_training_loss}
\end{figure}
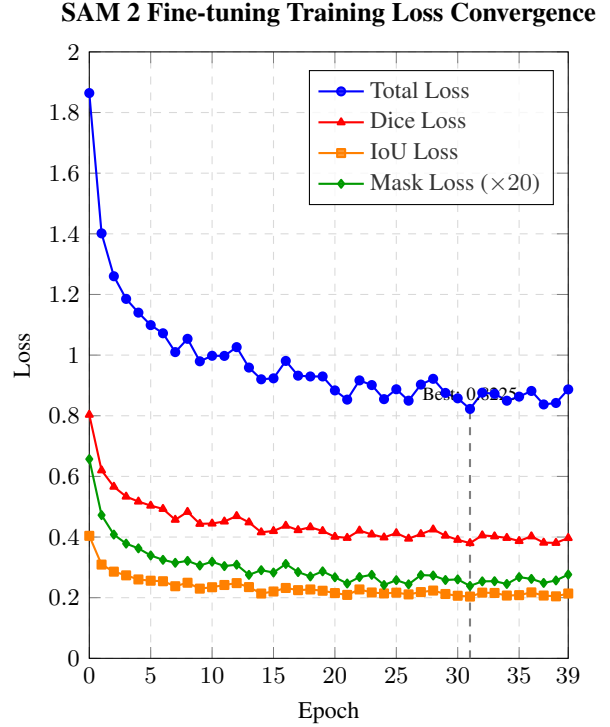

\subsubsection{Metrics}

\begin{figure}[htbp]
    \centering
    \begin{subfigure}[b]{0.48\columnwidth}
        \centering
        \includegraphics[width=\textwidth]{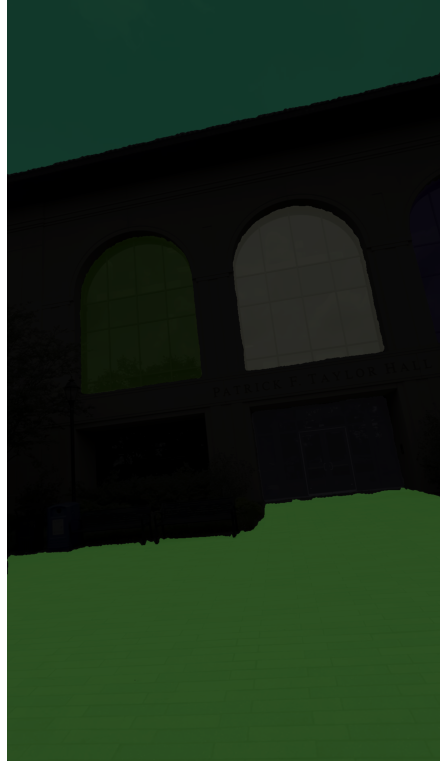}
        \caption{Fine-tuned SAM~2.1 on HBD dataset}
        \label{fig:pretrained}
    \end{subfigure}
    \hfill
    \begin{subfigure}[b]{0.48\columnwidth}
        \centering
        \includegraphics[width=\textwidth]{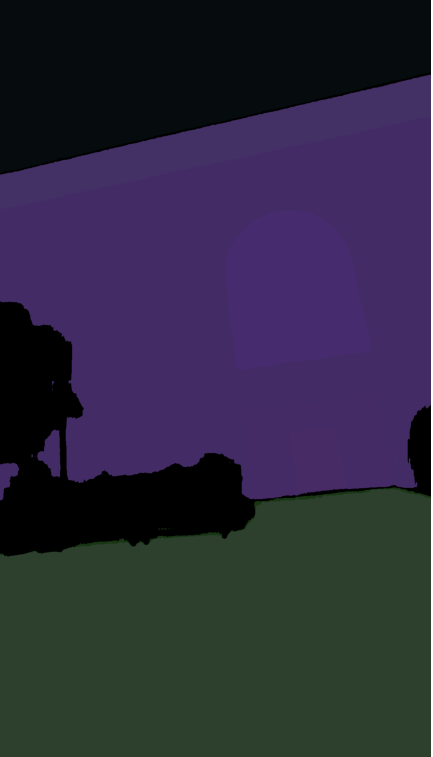}
        \caption{Pre-trained SAM~2.1 (Hiera-Large)}
        \label{fig:finetuned}
    \end{subfigure}
    \caption{Comparison of automatic mask generation on Patrick F. Taylor Hall. The fine-tuned model (a) produces more semantically meaningful segmentation of architectural elements, with clearer separation of the building facade, windows, and ground plane compared to the pre-trained model (b).}
    \label{fig:sam2_comparison}
\end{figure}

Table~\ref{tab:seg_compare} reports per-class IoU and Dice on the HBD test
set (349 images, 5{,}541 annotated instances). Fine-tuning SAM~2.1 on the
building-component dataset raises mean IoU from 49.9\% to 69.6\% (+19.8
points) and mean Dice from 59.3\% to 77.3\%. The improvement is consistent:
IoU increases for ten of the eleven classes, with the largest gains on
components that differ most from SAM~2's general-domain pre-training, namely
Curtain Wall (+35.5), Door (+34.7), Ceiling (+20.6), and Roof (+20.2).
Frequent structural classes also improve substantially (Wall +17.7, Window
+17.0). The only class that does not benefit is Opening ($-5.7$ IoU), which we
attribute to its ambiguous spatial extent---openings such as doorways and
passages lack a physical surface, and the fine-tuned model tends to merge them
with the surrounding wall. Wall remains the most difficult class even after
fine-tuning (53.0\% IoU), consistent with the challenge of segmenting large,
texture-poor surfaces from a single prompt. Roof and Lift are represented by
few instances (38 and 21) and their scores are therefore comparatively noisy.
All numbers use a single center-click prompt per instance following the
interactive protocol of SAM~2~\cite{ravi2024sam2}; the class label is taken
from the ground truth, so the metric measures mask quality per class rather
than class prediction.

\subsection{Baseline Comparison with SegFormer}

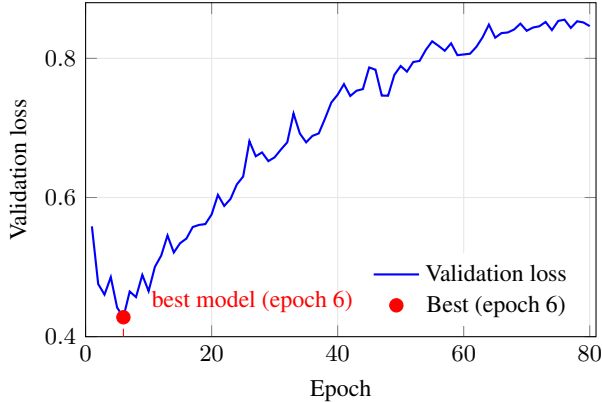
\begin{figure}[t]
\centering
\begin{tikzpicture}
\begin{axis}[
    width=\linewidth, height=6cm,
    xlabel={Epoch}, ylabel={Validation loss},
    xmin=0, xmax=81, ymin=0.40, ymax=0.88,
    grid=both, grid style={gray!20},
    tick label style={font=\small}, label style={font=\small},
    legend pos=south east, legend style={font=\small, draw=none},
]
\addplot[blue, thick] coordinates {
(1,0.5585)(2,0.4757)(3,0.4605)(4,0.4856)(5,0.4419)(6,0.4279)(7,0.4649)(8,0.4569)(9,0.4887)(10,0.4655)
(11,0.5004)(12,0.5166)(13,0.5454)(14,0.5210)(15,0.5344)(16,0.5410)(17,0.5576)(18,0.5606)(19,0.5618)(20,0.5757)
(21,0.6038)(22,0.5879)(23,0.5980)(24,0.6188)(25,0.6301)(26,0.6809)(27,0.6593)(28,0.6648)(29,0.6523)(30,0.6578)
(31,0.6690)(32,0.6793)(33,0.7207)(34,0.6920)(35,0.6793)(36,0.6886)(37,0.6921)(38,0.7136)(39,0.7366)(40,0.7478)
(41,0.7630)(42,0.7459)(43,0.7535)(44,0.7558)(45,0.7869)(46,0.7837)(47,0.7465)(48,0.7463)(49,0.7763)(50,0.7890)
(51,0.7809)(52,0.7948)(53,0.7963)(54,0.8123)(55,0.8245)(56,0.8180)(57,0.8109)(58,0.8216)(59,0.8046)(60,0.8056)
(61,0.8066)(62,0.8164)(63,0.8303)(64,0.8485)(65,0.8297)(66,0.8363)(67,0.8372)(68,0.8415)(69,0.8499)(70,0.8396)
(71,0.8443)(72,0.8461)(73,0.8523)(74,0.8406)(75,0.8538)(76,0.8556)(77,0.8437)(78,0.8533)(79,0.8516)(80,0.8463)
};
\addlegendentry{Validation loss}
\addplot[red, only marks, mark=*, mark size=2.5pt] coordinates {(6,0.4279)};
\addlegendentry{Best (epoch 6)}
\draw[red, dashed] (axis cs:6,0.40) -- (axis cs:6,0.4279);
\node[red, font=\small, anchor=west] at (axis cs:9,0.45) {best model (epoch 6)};
\end{axis}
\end{tikzpicture}
\caption{SegFormer validation loss per epoch during fine-tuning. The minimum is at
epoch~6 (marked); validation loss rises afterwards, so the epoch-6 checkpoint is
retained. The training loss (not shown) continues to fall to ${\sim}0.03$, indicating
overfitting.}
\label{fig:segformer_loss}
\end{figure}

Figure~\ref{fig:segformer_loss} shows the SegFormer validation loss during
fine-tuning. The training loss decreases steadily to approximately $0.03$, but the
validation loss reaches its minimum at epoch~6 ($0.43$) and rises thereafter,
indicating overfitting---expected given the capacity of the MiT-B3 encoder relative
to the size of the dataset. We therefore retain the best checkpoint (epoch~6)
selected by validation loss. On the test split this model attains $64.8\%$ mIoU and
$77.0\%$ mDice (Table~\ref{tab:seg_compare}). SegFormer is strongest on large,
contiguous surfaces (Wall $83.2\%$, Floor $87.0\%$, Ceiling $79.9\%$ IoU) and weaker
on small or infrequent components (Opening $38.3\%$, Roof $39.4\%$, Beam $44.5\%$,
Column $49.1\%$). Under the protocol caveats above, its mean IoU ($64.8\%$) exceeds
the pre-trained SAM~2.1 baseline ($49.9\%$) but remains below our fine-tuned SAM~2.1
($69.6\%$).

Table~\ref{tab:seg_compare} reports per-class IoU and Dice on the HBD test set.
Under an identical prompting protocol, fine-tuning improves SAM~2.1 substantially
over the pre-trained model: mean IoU rises from 49.9\% to 69.6\% (+19.8 points)
and mean Dice from 59.3\% to 77.3\%, with gains on ten of the eleven classes (all
except Opening). This is our controlled comparison, as both SAM variants receive
the same center-click prompts and the same ground-truth class assignment.

As a reference point, a fully-automatic SegFormer trained on the same data reaches
64.8\% mIoU and 77.0\% mDice. This is not a strict head-to-head: SegFormer predicts
both the mask and the class with no prompt and is scored per pixel, whereas SAM~2.1
is prompted from the ground truth and scored per object. The two are therefore
evaluated under different protocols and should be read as complementary references
rather than directly comparable numbers.

The per-class breakdown shows complementary strengths. SegFormer is stronger on
large, contiguous surfaces (Wall 83.2 vs 53.0 IoU; Ceiling 79.9 vs 65.2), which a
per-pixel semantic segmenter covers naturally, while prompted SAM~2.1 is stronger
on discrete or less frequent components (Column 71.7 vs 49.1; Curtain Wall 68.3 vs
53.2; Roof 63.0 vs 39.4; Opening 58.7 vs 38.3), where a single click isolates the
object cleanly. Window, Floor, and Lift are comparable. Roof and Lift have few
instances (38 and 21), so their scores are noisier.

\begin{table}[tp]
\centering
\small
\setlength{\tabcolsep}{5pt}
\caption{Per-class segmentation accuracy on the HBD test set (349 images, 5{,}541 annotated instances).
Pre-trained SAM~2.1 (Pre.) and our fine-tuned SAM~2.1 (FT) are both prompted by a single center click
derived from the ground-truth mask and scored per object. SegFormer (Seg.) is a fully-automatic semantic
segmenter scored per pixel, reported for reference. \textbf{Bold} marks the better of the two matched SAM
variants; SegFormer uses a different protocol and is not a strict head-to-head. Scores are macro-averaged over classes.}
\label{tab:seg_compare}
\begin{tabular}{l r ccc ccc}
\toprule
 & & \multicolumn{3}{c}{IoU (\%)} & \multicolumn{3}{c}{Dice (\%)} \\
\cmidrule(lr){3-5}\cmidrule(lr){6-8}
Class & \#Obj & Pre. & FT & Seg.$^{\dagger}$ & Pre. & FT & Seg.$^{\dagger}$ \\
\midrule
Wall        & 1679 & 35.3 & \textbf{53.0} & 83.2 & 44.6 & \textbf{62.4} & 90.8 \\
Window      & 1331 & 54.3 & \textbf{71.3} & 70.6 & 63.0 & \textbf{78.8} & 82.8 \\
Ceiling     & 673  & 44.6 & \textbf{65.2} & 79.9 & 54.5 & \textbf{74.0} & 88.8 \\
Door        & 474  & 51.8 & \textbf{86.5} & 83.1 & 61.3 & \textbf{91.4} & 90.7 \\
Column      & 383  & 62.6 & \textbf{71.7} & 49.1 & 71.0 & \textbf{80.6} & 65.9 \\
Beam        & 324  & 41.8 & \textbf{54.7} & 44.5 & 50.6 & \textbf{64.5} & 61.6 \\
Floor       & 309  & 70.1 & \textbf{86.4} & 87.0 & 77.1 & \textbf{90.8} & 93.1 \\
Opening     & 219  & \textbf{64.4} & 58.7 & 38.3 & \textbf{74.6} & 66.6 & 55.4 \\
Curtain Wall & 90   & 32.8 & \textbf{68.3} & 53.2 & 41.7 & \textbf{76.0} & 69.5 \\
Roof        & 38   & 42.8 & \textbf{63.0} & 39.4 & 51.7 & \textbf{73.0} & 56.6 \\
Lift        & 21   & 47.9 & \textbf{86.9} & 84.6 & 61.6 & \textbf{92.2} & 91.7 \\
\midrule
\textbf{Mean} & 5541 & 49.9 & \textbf{69.6} & 64.8 & 59.3 & \textbf{77.3} & 77.0 \\
\bottomrule
\end{tabular}

\vspace{2pt}
{\footnotesize $^{\dagger}$SegFormer is fully automatic (predicts its own class, no prompt) and scored per pixel; shown for reference, not a strict head-to-head.}
\end{table}

\begin{figure*}[tp]
\centering
\includegraphics[width=0.95\textwidth]{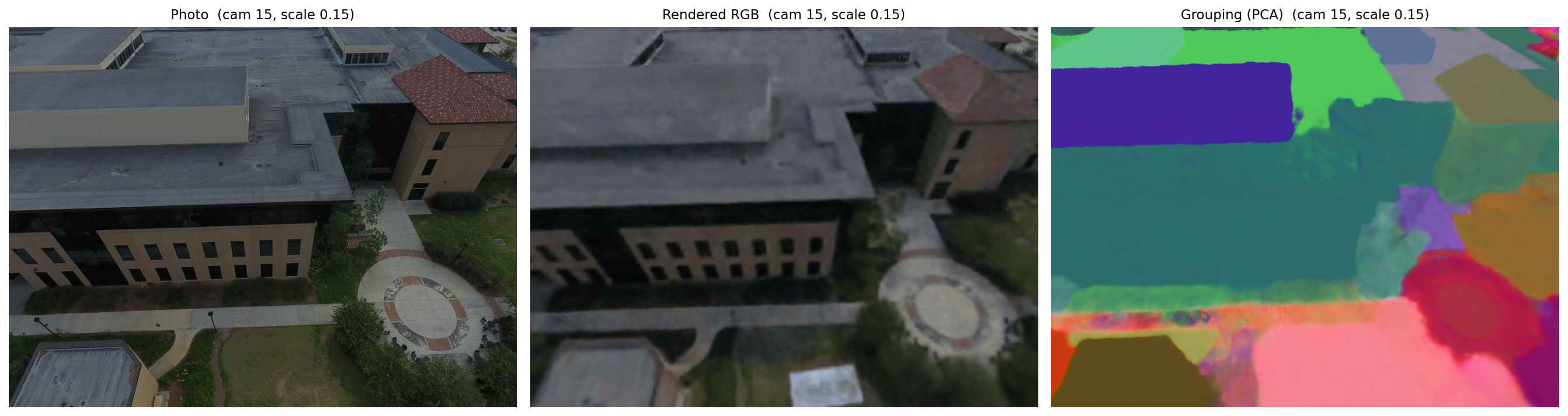}\\[1pt]
{\small (a) Pre-trained SAM~2.1 masks --- grouping scale 0.15}\\[5pt]
\includegraphics[width=0.95\textwidth]{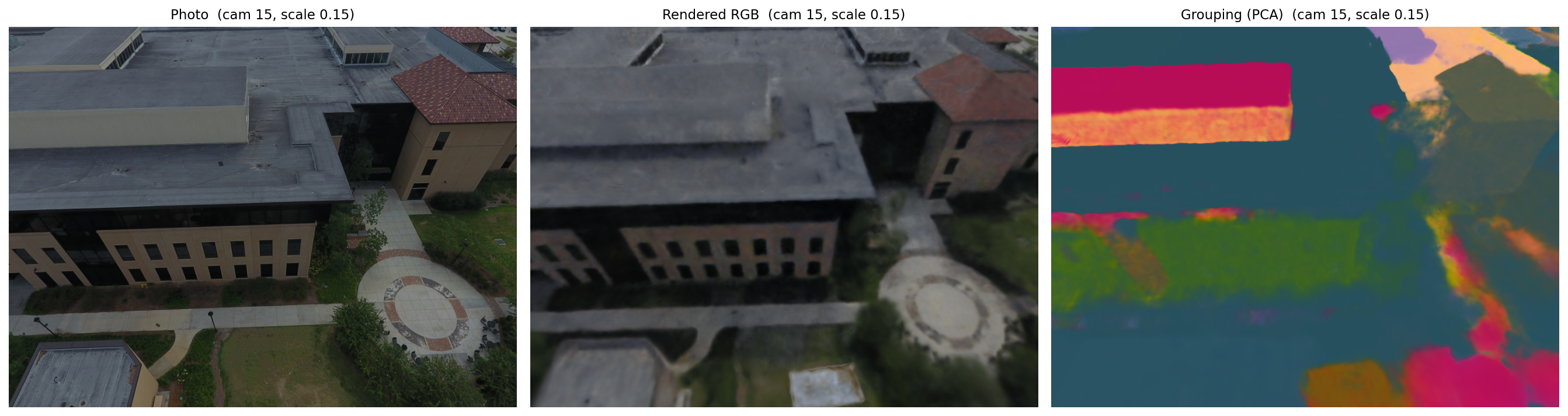}\\[1pt]
{\small (b) Fine-tuned SAM~2.1 masks --- grouping scale 0.15}\\[5pt]
\includegraphics[width=0.95\textwidth]{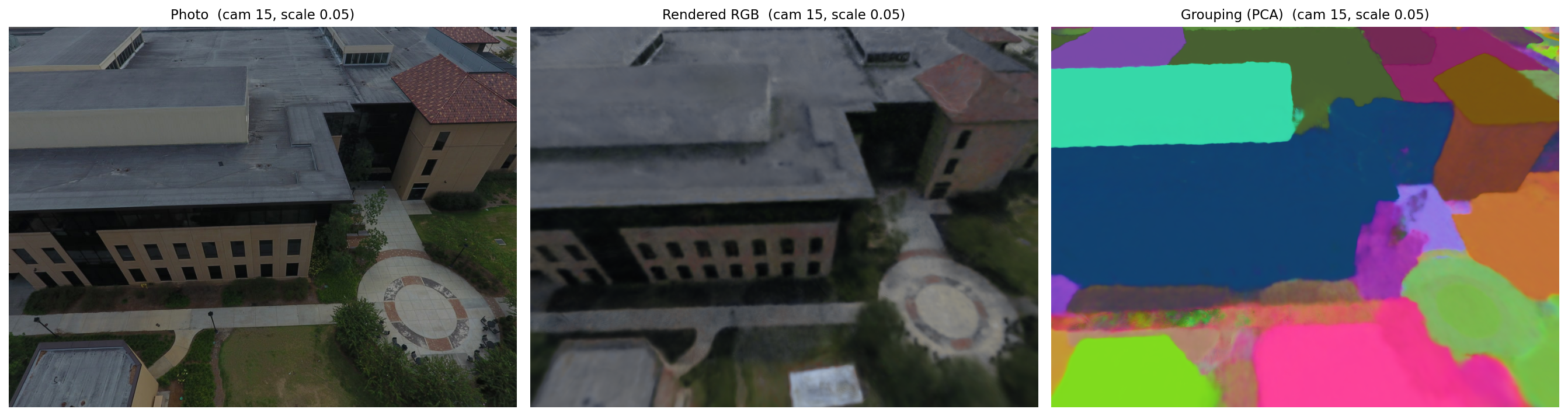}\\[1pt]
{\small (c) Pre-trained SAM~2.1 masks --- grouping scale 0.05}\\[5pt]
\includegraphics[width=0.95\textwidth]{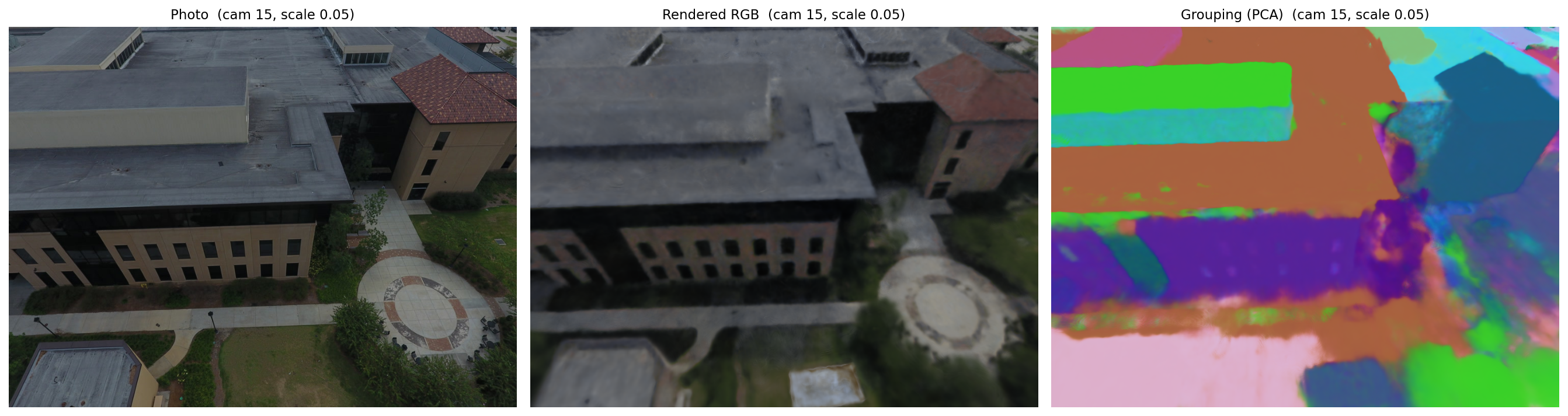}\\[1pt]
{\small (d) Fine-tuned SAM~2.1 masks --- grouping scale 0.05}
\caption{Qualitative 3D grouping on the Patrick F.\ Taylor Hall drone scene, from a fixed
viewpoint (camera 15), at two grouping scales. Each row shows, left to right, the input
photograph, the reconstructed RGB, and the GARField grouping field visualized by
projecting its per-point features to RGB with PCA. Rows (a,b) use grouping scale 0.15 and
rows (c,d) use scale 0.05; (a,c) use pretrained SAM~2.1 masks and (b,d) use our fine-tuned
SAM~2.1 masks. Both models are trained with an identical configuration and differ only in
the SAM~2.1 checkpoint. Because the PCA projection is computed independently per image,
absolute colors are not comparable across rows; we compare the spatial coherence of the
groups.}
\label{fig:grouping_qual}
\end{figure*}

\subsection{3D Grouping Evaluation}\label{sec:eval}
\paragraph{Scene and data.} We evaluate on drone imagery of Patrick F. Taylor Hall at Louisiana State University (the ``PFTdrone'' scene), consisting of 1250 images at [4000 × 3000] resolution captured by a drone. Camera poses and a sparse point cloud were recovered with COLMAP structure-from-motion~\cite{schoenberger2016sfm}.
\paragraph{Checkpoints.} The mask generator is SAM~2.1 (Hiera-L). For the baseline condition we use the public pretrained checkpoint (\texttt{sam2.1\_hiera\_large.pt}); for the fine-tuned condition we use our building-component fine-tuned checkpoint.
\paragraph{GARField training.} We build the radiance and grouping fields with nerfstudio~\cite{nerfstudio} using GARField (\cite{garfield2024}), trained for 30{,}000 steps on a single NVIDIA GeForce RTX~4070~Ti (12~GB) GPU, with an Intel Core~i9-13900K CPU and 128~GB of RAM. The PFTdrone scene contains 1{,}250 drone images (1{,}093 for training, 157 for evaluation), processed at a downscale factor of~4 (effective resolution [1000 × 750], from an original [4000 × 3000]). Camera poses were recovered with COLMAP~\cite{schoenberger2016sfm} using spatial matching with the images’ EXIF information. The pretrained and fine-tuned conditions used an identical configuration and differed only in the SAM~2.1 checkpoint used to generate the grouping masks; the cached masks were regenerated between runs.

Figure \ref{fig:grouping_qual} compares the 3D grouping produced by GARField when it is
supervised with masks from the pretrained versus the fine-tuned SAM~2.1, on our Patrick
F.\ Taylor Hall drone scene. For a fixed drone
viewpoint, we show the input photograph, the reconstructed RGB, and the grouping field
visualized by projecting its per-point features to RGB with PCA at a fixed grouping
scale. Because the PCA projection is computed independently for each image, the absolute
colors are not comparable between the two models; we therefore compare the spatial
coherence of the groups rather than their colors.
In the example shown, the fine-tuned model produces more meaningful grouping of architectural components, particularly around the roof and facade regions, whereas the pretrained model merges portions of these components with neighboring surfaces. The differences are more apparent at the coarser grouping scale, while they become less pronounced at finer scales. Because this experiment is qualitative, we treat these results as illustrative. These results suggest that domain adaptation not only produces more meaningful groupings in 3D but also corrects previously merged regions.

\section{Conclusion}

Semantic segmentation of building components in 2D images is a vital part of image-based 3D reconstruction of building models. In this regard, domain adaptation of a vision foundation model is required to enable 3D hierarchical grouping segmentation of buildings from images of specific building components, such as facades or windows. This study showed that fine-tuning SAM 2.1 as an automatic mask generator can enhance the segmentation of building components when they are grouped by Group Anything with Radiance Fields (GARField). Fine-tuning was performed on a large building dataset, achieving a significant 55.9\% decrease in loss and better segmentation of some object classes compared with SegFormer as a baseline. As future work, further quantitative evaluation, alongside our qualitative evaluation, can be conducted to demonstrate grouping performance when segmentation models are used for 3D reconstruction in the building domain.

\section*{Large Language Models}
A large language model (Claude, Anthropic) was used in preparing this
work. It assisted with editing and language polishing of the manuscript
text (including the abstract), and with drafting and debugging evaluation
code: the SAM~2.1 per-class IoU/Dice evaluation (including the
deterministic center-point prompting rule), the SegFormer training and
evaluation scripts, and the rendered-2D IoU evaluation used to compare
GARField's 3D grouping under pretrained versus fine-tuned SAM~2.1 masks.
All experimental design, parameter choices, and analysis of the results
are the authors' own; the model was not used to generate or select any
reported numerical result, and all text was reviewed and edited by the
authors for correctness.

{
    \small
    \bibliographystyle{ieeenat_fullname}
    \bibliography{main}
}

\end{document}